\pdfoutput=1
\documentclass[letterpaper]{article} %
\usepackage[preprint]{aaai2027}  %
\usepackage[hyphens]{url}  %
\usepackage{graphicx} %
\usepackage{natbib}  %
\usepackage{caption} %
\usepackage{amsmath}
\usepackage{booktabs}
\usepackage{xcolor}
\usepackage{multirow}

\title{Report: Progressive Disclosure of Agent Skills}

\author{
    Guilin Zhang,
    Kai Zhao,\footnote{Corresponding author: kai.zhao@workday.com}
    Priyanka Mudgal,
    Waleed Ammar,
    Xiquan Cui,
    Xu Chu,
    Alet Blanken
}
\affiliations{
    Workday AI Research
}

\begin{document}

\maketitle

\begin{figure*}[t]
\centering
\includegraphics[width=\textwidth]{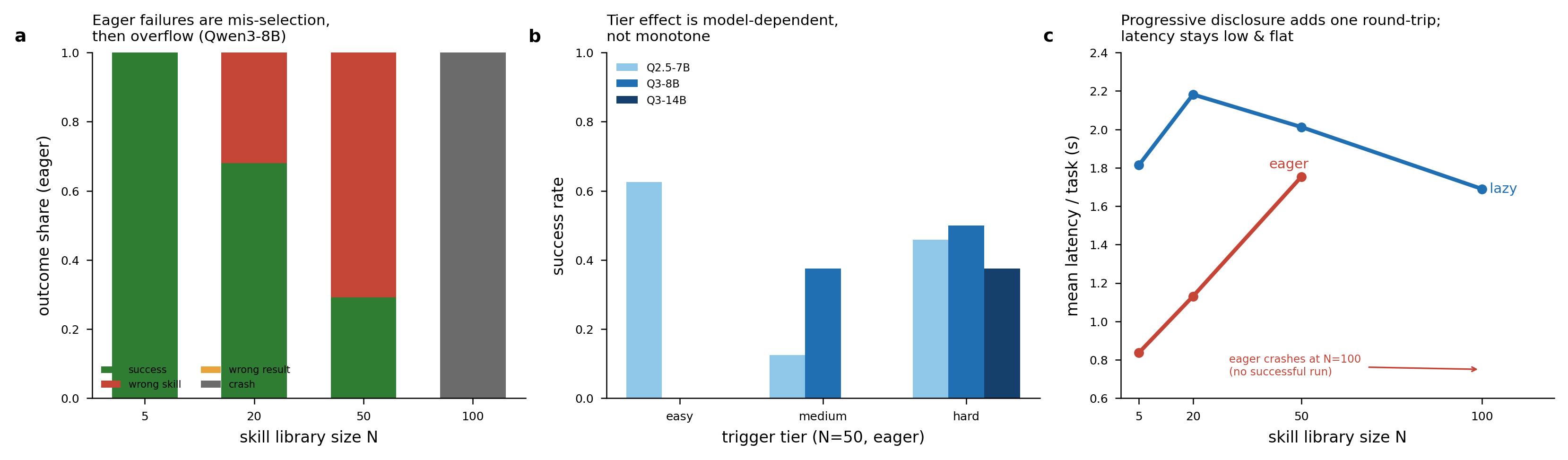}
\caption{(a) Under the eager loading regime, skill-retrieval quality degrades as the skills library size $N$ grows (Qwen3-8B). 
(b) Under the eager loading regime, the larger core LLM (Qwen3-14B) is reliable against easy/medium-difficulty skill distractors, but susceptible to hard distractors. 
(c) Under the progressive disclosure regime, the LLM core is invoked more often which increases latency, especially for smaller values of the library size $N$.}
\label{fig:Asupp}
\end{figure*}

\begin{abstract}
Users of Workday's deployed LLM-based agents often request features which can be addressed by defining named procedures, also known as skills, in the LLM context, effectively augmenting agents' capabilities. However, as an agent's skills library grows in size, so does the agent's operational cost. Progressive disclosure (lazy-loading) of skills as needed may reduce operational costs, but its impact on overall latency and skill-retrieval quality remains unclear. In this report, we investigate the impact empirically and find that progressive disclosure improves skill-retrieval quality but marginally degrades overall latency.
\end{abstract}

\section{Agent Skills in Production}
Workday\footnote{\url{https://www.workday.com/}} is a cloud-based enterprise platform for managing people, money and agents, serving over 65\% of Fortune 500 enterprises.
As of August 2026, over 5,500 of our customers are using one or more AI agents deployed in the Workday platform, a 35\% increase compared to the previous quarter.

\paragraph{Skills augment agentic capabilities.}
As more customers adopt our deployed LLM-based agents, it is necessary to adapt their agentic capabilities in response to feature requests and bug reports.
A simple and effective approach for enhancing agentic capabilities is to define specialized instructions detailing how a particular feature request or bug report may be addressed.
Modern agent frameworks, such as LangChain, package those instructions as \emph{skills}.
Skills package domain expertise, such as workflows, best practices, scripts, reference docs, and templates, into reusable directories.
Each skill has a name, a natural-language description, an input/output signature, and often a worked example or usage notes.

\paragraph{Eager loading of skills does not scale.}
In order to enable an agent's underlying large language model (LLM) to faithfully follow the instructions in a skill's definition, the `eager loading' skill management regime fully discloses the skill in the LLM prompt context.
However, as an agent's skills library grows in size, we observe a notable increase in its operational costs, due to the number of tokens used to disclose the skills library in the LLM context, which prompted us to explore alternative regimes for managing agent skills.

\paragraph{Progressive disclosure offers a viable solution.}
Lazy loading of skill definitions as needed, also known as progressive disclosure, is an attractive alternative, commonly adopted in agentic frameworks such as LangChain.
When progressive disclosure is enabled, agents incrementally disclose skills in the LLM context, pulling in more details only as a task calls for it.

In this report, we investigate the tradeoffs of activating progressive disclosure of skills in the agent harness, measuring how it impacts an agent's overall token-based operational cost, latency and skill-retrieval quality, as the library size grows in a controlled setup.
Next, we describe our implementation for progressive disclosure.

\section{Agent Skills Management in The Harness}
A minimal agent consists of an LLM core and its harness.\footnote{It is not uncommon for one agent to utilize multiple LLMs for different subtasks.}
The harness interfaces with the agent's environment, determines when to invoke the LLM, manages the skills library, prepares the context (and prompt) before invoking the LLM, and processes LLM outputs.

\paragraph{Curating a skills library.}
An agent's skills library consists of $N$ structured skills.
According to the open standard format for Agent Skills, each skill is a directory containing, at minimum, a \texttt{SKILL.md} file starting with the skill's frontmatter (a \emph{name} field and a \emph{description} field in YAML), followed by the skill's \emph{body} content which contains the operative content in Markdown, e.g., step-by-step instructions, input-output examples and edge cases.\footnote{The frontmatter may also include optional fields such as \emph{license}, \emph{compatibility} and \emph{metadata}. See \url{https://agentskills.io} for more details on the Agent Skills open standard.}
Instead of manually creating all skill content from scratch, coding agents such as Claude Code can be used to create and modify specialized skills and it is not uncommon for agent developers to share the skills they define in version-controlled repositories in order for them to be reused in multiple agents and by other developers.\footnote{See \url{https://officialskills.sh/} for example skill repositories.}

\paragraph{Two regimes for operating a skills library.} 
A naïve regime an agent harness may use is to port all details of all $N$ skills in its library to the LLM context, which we call `eager loading' of skills.
As the number of skills increase, eager loading substantially increases the number of tokens needed to disclose skills in the LLM context, which impacts the agent's quality, reliability and cost.
An alternative regime loads skills progressively, disclosing minimal information about available skills in the LLM context early on and adding more details about the most relevant skills in subsequent LLM invocations only when a task calls for it, also known as `progressive disclosure'.
In this report, we contrast eager loading with a regime that loads the frontmatter of all $N$ skills in the LLM context and prompts the LLM core to emit a structured command which determines which skill is most relevant for the task at hand, e.g., \texttt{\{"action":"load\_skill","name":"pptx"\}}.
The harness then loads the \emph{body} content of the chosen skill to the LLM context in subsequent LLM invocations.\footnote{Note that this implementation requires additional invocations to the LLM core, but reduces the number of tokens used in the LLM context.}

\begin{figure*}[t]
\centering
\includegraphics[width=\textwidth]{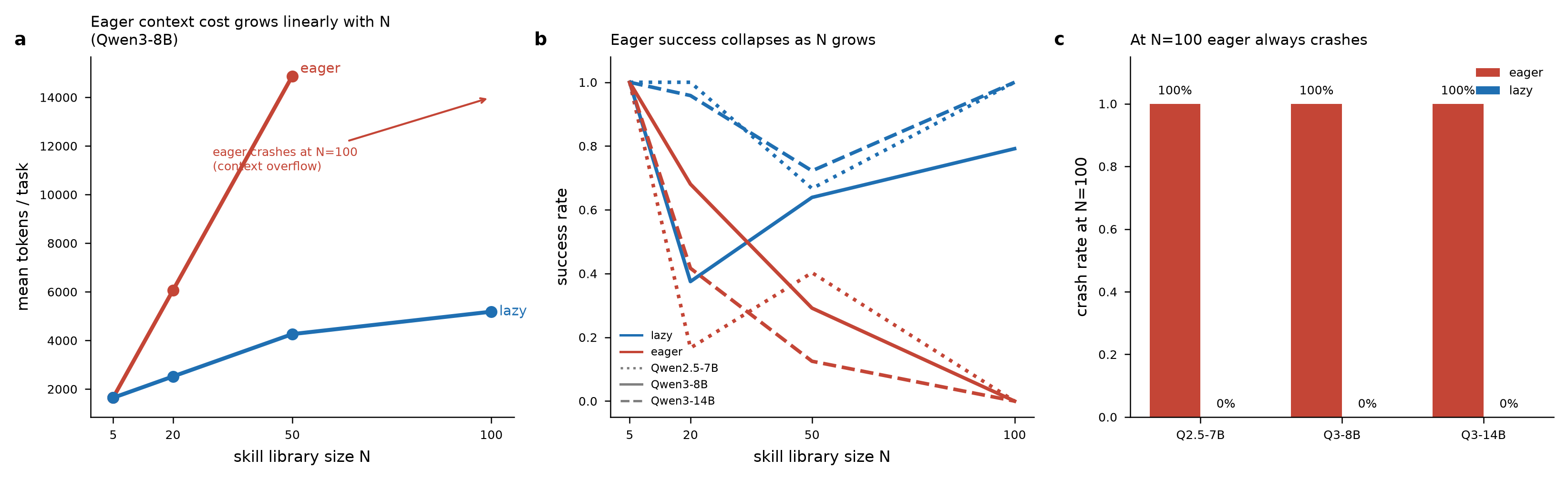}
\caption{(a) Prompt token usage grows faster in the eager loading regime. 
(b) Skill-retrieval quality degrades faster in the eager loading regime, as $N$ grows. 
(c) The rate at which each skill management regime overflows the maximum context size at library size $N{=}100$.}
\label{fig:A}
\end{figure*}

\section{Experimental Setup}
We estimate the impact of progressive disclosure by measuring skill-retrieval success rate, total tokens used and overall latency in seconds.

\paragraph{Skill retrieval.}
The impact of progressive disclosure on an agent's quality is mediated by its ability to select relevant skills, which we quantify using an explicit skill-retrieval task.
We augment the \emph{body} content of each skill with a unique `activation code' identifier, produced by a random number generator.
Each instance of the task names a domain intent (e.g., travel) and asks the agent to return the most relevant activation code in a structured response which takes the form: \texttt{RESULT[<skill-name>]:\,<activation-code>}. 
The agent harness delegates the task to the LLM core, which is only able to return the correct activation code if the relevant skill's body is loaded in its context.
We define a taxonomy of outcomes: 
\begin{itemize}
    \item \emph{success} (skill name and activation code are both correct), 
    \item \emph{wrong\_name} (skill name is incorrect), 
    \item \emph{wrong\_code} (skill name is correct, but the activation code is wrong), and 
    \item \emph{crash} (context overflow or malformed output). 
\end{itemize}
We experiment with four library sizes: $N \in \{5, 20, 50, 100\}$, and develop a core set of $24$ task instances based on $5$ relevant skills.
For $N>5$, we pad the skills library with distractor skills with three difficulty tiers that vary the directness of the trigger phrasing, drawn from a generator spanning $12$ domain families.
We use three different seeds for each task, and report means over $72$ rollouts for each unique combination of library size $N$ and skill management regime (i.e., eager loading vs. progressive disclosure).

\paragraph{LLM cores.} We experiment with three LLM cores: Qwen2.5-7B-Instruct \cite{qwen25}, Qwen3-8B and Qwen3-14B \cite{qwen3}, with greedy decoding and a $32$k context window.\footnote{Qwen3 models run in non-thinking mode.}
For each LLM core, we run $576$ agent rollouts, which correspond to $4$ library sizes $\times\ 2$ skill management regimes $\times\ 24$ task instances $\times\ 3$ seeds.
We use vLLM \cite{kwon2023vllm} on a single NVIDIA L40S (48\,GB) to serve all LLM cores. 
Every core LLM invocation passes through a wrapper that records prompt tokens count, completion tokens count, and wall-clock latency from the server's usage accounting.\footnote{We report per-task totals summed over all calls in a rollout, so the extra \texttt{load\_skill} LLM invocations used when progressive disclosure is enabled are accounted for.}

\section{Key Findings}

\paragraph{1. Progressive disclosure reduces operational cost and improves reliability.}
When eager loading is enabled, skill definitions increasingly dominate the context as $N$ grows, eventually breaking the agent's reliability due to exceeding the maximum allowed prompt size, indicated by \emph{crash} in all rollouts with $N=100$ in Table~\ref{tab:paperA} (Eager tok) and Figure~\ref{fig:A} (c).
Table~\ref{tab:paperA} (Tok $\downarrow$) shows the percentage of tokens saved when progressive disclosure is enabled, demonstrating savings across all values of $N$, as intended, as a direct result of only disclosing the full \emph{body} of the most relevant skill in progressive disclosure.
Progressive disclosure demonstrates increasingly bigger token savings as $N$ grows, up to $81.7\%$ of tokens originally used in the eager loading regime.
Figure~\ref{fig:A} (a) visually demonstrates how token usage grows as a function of library size in both regimes.
With the prevalent token-based pricing of LLMs, these savings translate to substantial savings of agents' operational cost at scale.

\paragraph{2. Progressive disclosure may increase overall latency.}
As we described earlier, the progressive disclosure regime for skill management entails an additional LLM invocation dedicated to identifying relevant skills, which often increases the agent's mean overall latency, as illustrated in Figure~\ref{fig:Asupp} (c).
For example, using progressive disclosure instead of eager loading corresponds to a $15\%$ increase in the agent's overall latency (from $1.75$ to $2.01$ seconds) when using Qwen3-8B with a library of size $N{=}50$.\footnote{We note that the delta reduces as $N$ grows, which may reverse this result for a larger prompt token budget. 
Depending on task complexity, an agent may amortize the latency associated with extra LLM invocations in progressive disclosure over more subtasks which may also reverse this result.}

\paragraph{3. Progressive disclosure improves skill-retrieval quality, on average.}
While progressive disclosure is the undisputed winner on average skill retrieval quality, neither regime consistently outperforms the other, as shown in Figure~\ref{fig:A} (b).
Progressive disclosure either matches or outperforms eager loading on average skill-retrieval quality in all but one experimental setup, as shown in Table~\ref{tab:paperA} (Skill retrieval).
As the library size $N$ grows, distractor skills dominate the LLM context, and the skill-retrieval quality for the eager loading regime falls sharply (e.g., from $1.00$ at $N{=}5$ to $0.125$ at $N{=}50$ for Qwen3-14B), exposing retrieval of the wrong skill name as the dominant failure as shown in Figure~\ref{fig:Asupp} (a).\footnote{\citet{levy2024sametask} also report degradation in reasoning quality as the LLM prompt size increases.}
We note that the largest LLM core (Qwen3-14B) is resilient against easy/medium distractors but remains susceptible to hard distractors, as shown in Figure~\ref{fig:Asupp} (b).
This finding emphasizes the importance of developing skill management regimes which do not only optimize for token usage but also skill retrieval quality.
\begin{table*}[h]
\centering
\small
\begin{tabular}{l|l|rrr|rr|rr}
\toprule
 &  & \multicolumn{3}{c|}{\textbf{Token usage}} & \multicolumn{2}{c|}{\textbf{Skill retrieval}} & \multicolumn{2}{c}{\textbf{Reliability}} \\
\textbf{LLM core} & $N$ & EL tok & PD tok & Tok $\downarrow$ & EL succ & PD succ & EL crash & PD crash \\
\midrule
\multirow{4}{*}{Qwen2.5-7B} & 5   & 1648  & 1215 & 26.3\% & 1.00 & 1.00 & 0.00 & 0.00 \\
 & 20  & 6057  & 1943 & 67.9\% & 0.17 & 1.00 & 0.00 & 0.00 \\
 & 50  & 14871 & 3481 & 76.6\% & 0.40 & 0.67 & 0.00 & 0.00 \\
 & 100 & \textit{crash} & 6223 & --- & 0.00 & 1.00 & 1.00 & 0.00 \\
\midrule
\multirow{4}{*}{Qwen3-8B} & 5   & 1652  & 1644 & 0.5\%  & 1.00 & 1.00 & 0.00 & 0.00 \\
 & 20  & 6060  & 2519 & 58.4\% & 0.68 & 0.38 & 0.00 & 0.00 \\
 & 50  & 14876 & 4262 & 71.3\% & 0.29 & 0.64 & 0.00 & 0.00 \\
 & 100 & \textit{crash} & 5184 & --- & 0.00 & 0.79 & 1.00 & 0.00 \\
\midrule
\multirow{4}{*}{Qwen3-14B} & 5   & 1652  & 973  & 41.1\% & 1.00 & 1.00 & 0.00 & 0.00 \\
 & 20  & 6061  & 1553 & 74.4\% & 0.42 & 0.96 & 0.00 & 0.00 \\
 & 50  & 14875 & 2719 & 81.7\% & 0.13 & 0.72 & 0.00 & 0.00 \\
 & 100 & \textit{crash} & 4661 & --- & 0.00 & 1.00 & 1.00 & 0.00 \\
\bottomrule
\end{tabular}
\caption{Eager loading (EL) vs.\ progressive disclosure (PD) results as a function of library size $N$. 
Tokens are per-task means over $72$ rollouts.
\emph{crash} denotes context overflow. 
Token reductions at $N\ge 20$ are significant at $p<10^{-25}$, based on one-sided Mann--Whitney $U$ test on per-rollout totals.}
\label{tab:paperA}
\end{table*}

\section{Conclusion}
Skills have emerged as an effective way of augmenting agentic capabilities in response to increasing adoption of production agents.
Progressive disclosure declares the frontmatter (a skill's name and brief description) of all skills available to an agent in the initial LLM context then incrementally loads the full definition of relevant skills as needed, instead of naïvely loading the full definitions of all skills.
In this report, we found that progressive disclosure of agent skills reduces token usage and improve reliability as well as skill-retrieval quality, at the expense of greater latency.

\section{Open Questions}
\label{sec:limA}
\paragraph{How many skills do we need?} 
The experimental setup discussed earlier limits skill retrieval to one skill for each task.
In practice, the agent has no prior knowledge about the number of skills needed for a given task.
How do we balance the verbosity of skill definitions with the complexity of the task at hand remains an open research question in agent skill management with critical consequences.

\paragraph{Which skill files do we need?}
The experimental setup discussed earlier limits the definition of each skill in the library to its \texttt{SKILL.md} file.
In practice, a skill's directory may also contain additional files (e.g., scripts, references or assets) which would take up even more room in the LLM context.
What strategies do we use to determine which files to load?
How do we evaluate the efficacy of different strategies in practice?

\paragraph{Which skills are safe to include?}
With increased adoption of skill libraries, agent developers may accidentally include serious vulnerabilities hidden in skill definitions.
How do we effectively monitor an agent's risk exposure as a result of including a skill in its library?

\bibliography{refs}

\end{document}